\pdfoutput=1
\documentclass[11pt]{article}

\usepackage[]{ACL2023}

\usepackage{times}
\usepackage{latexsym}

\usepackage[T1]{fontenc}
\usepackage[utf8]{inputenc}

\usepackage{microtype}

\usepackage{inconsolata}

\usepackage{url}
\usepackage{subfigure}
\usepackage{multirow}
\usepackage{enumitem}
\usepackage{stmaryrd}
\usepackage{array}
\usepackage{threeparttable}
\usepackage{blindtext}
\usepackage{amssymb}

\usepackage[ruled,vlined,linesnumbered,longend]{algorithm2e}
\usepackage{svg}
\usepackage{placeins}

\usepackage{soul}
\usepackage[utf8]{inputenc}
\usepackage{graphicx}
\usepackage{amsmath}
\usepackage{booktabs}
\usepackage{lipsum}
\usepackage{listings}
\usepackage{xcolor, soul}
\sethlcolor{blue!10}
\usepackage{hyperref}
\newcommand{\hide}[1]{}

\newcommand{\korc}{\textsc{KoRC}\xspace}
\newcommand{\korct}{\textsc{KoRC-T}\xspace}
\newcommand{\korch}{\textsc{KoRC-H}\xspace}
\newcommand{\korcl}{\textsc{KoRC-L}\xspace}

\usepackage{amsmath,amsfonts,bm}

\def\eqref#1{equation~\ref{#1}}
\def\1{\bm{1}}

\def\rva{{\mathbf{a}}}

\DeclareMathAlphabet{\mathsfit}{\encodingdefault}{\sfdefault}{m}{sl}
\SetMathAlphabet{\mathsfit}{bold}{\encodingdefault}{\sfdefault}{bx}{n}

\def\gK{{\mathcal{K}}}

\title{
\korc: Knowledge oriented Reading Comprehension \\
Benchmark for Deep Text Understanding
}

\newcommand*{\email}[1]{\texttt{#1}}

\author{
Zijun Yao$^{1,2}$\thanks{\quad Yao and Liu contributes equally to KoRC. Work is done when Liu is an intern at Zhipu.AI.} \quad 
Yantao Liu$^{3,4*}$ \quad
Xin Lv$^{1,2}$ \\
{\bf Shulin Cao}$^{1,2}$ \quad 
{\bf Jifan Yu}$^{1,2}$ \quad 
{\bf Lei Hou}$^{1,2}$ \quad 
{\bf Juanzi Li}$^{1,2}$\thanks{\quad Corresponding author.} \quad  \\
Department of Computer Science and Technology, \\
$^1$BNRist; $^2$KIRC, Institute for Artificial Intelligence \\
Tsinghua University, Beijing 100084, China \\
$^3$University of Chinese Academy of Sciences $^4$Zhipu.AI \\
\email{yaozj20@mails.tsinghua.edu.cn}, 
\email{\{houlei,lijuanzi\}@tsinghua.edu.cn}
}

\begin{document}
\maketitle
\begin{abstract}

Deep text understanding, which requires the connections between a given document and prior knowledge beyond its text, has been highlighted by many benchmarks in recent years. 
% However, these benchmarks are usually challenged by two main limitations, i.e., limited knowledge coverage and narrow answer space. 
However, these benchmarks have encountered two major limitations.
% are usually challenged by two main limitations. 
On the one hand, most of them require human annotation of knowledge, which leads to limited knowledge coverage. On the other hand, they usually use choices or spans in the texts as the answers, which results in narrow answer space.
To overcome these limitations, we build a new challenging benchmark named \korc in this paper. Compared with previous benchmarks, \korc has two advantages, \textit{i.e.,} broad knowledge coverage and flexible answer format. Specifically, we utilize massive knowledge bases to guide annotators or large language models (LLMs) to construct knowledgable questions. Moreover, we use labels in knowledge bases rather than spans or choices as the final answers.
We test state-of-the-art models on KoRC and the experimental results show that the strongest baseline only achieves $68.3\%$ and $30.0\%$ F1 measure in the in-distribution and out-of-distribution test set, respectively.
These results indicate that deep text understanding is still an unsolved challenge.
The benchmark dataset, leaderboard, and baseline methods are released in \url{https://github.com/THU-KEG/KoRC}.
% We will release our dataset and baseline methods upon acceptance.

% In this paper, we present KoRC, a large-scale \textbf{K}nowledge \textbf{o}riented \textbf{R}eading \textbf{C}omprehension dataset of $3,291$ documents and $34,383$ questions with broad knowledge coverage and flexible answer format.
% Different from previous machine reading comprehension datasets, KoRC answers questions by multi-hop reasoning.
% It starts from information in the document and ends up finding answer entities in background knowledge bases. 
% To construct KoRC, we design methods for template-based question generation and human engaged question annotation, as well as investigate the means to instruct large language models (LLMs) in order to generate high-quality questions. 
% We test state-of-the-art models on KoRC, which are capable to make connections with external knowledge, including the most recently released LLMs.
% Experimental results show that the most strong baseline only achieves 61.4\% and 26.4\% F1 measure in the IID and OOD test set, respectively.
% Furthermore, we find that LLMs fail to correctly answer question even generated by themselves.
% The test results of KoRC indicates that deep text understanding is still an unsolved challenge.
% We will release our dataset and baseline methods upon acceptance.

\end{abstract}
\section{Introduction}

% \textit{``Reading between the lines''}, which is to make inference beyond what is overtly stated in the text, is one of the fundamental requirements of deep text understanding.
% % It requires the reader to link information from the text with relevant background knowledge and making inference beyond what is overtly stated.
% It requires the skilled readers to integrate text information with its relevant background (prior) knowledge~\citep{gough1986decoding,castles2018ending,smith2021role}, and has been the long-pursuing goal in machine reading comprehension for decades~\citep{mccarthy1976example,norvig1987unified,huang2019cosmos}.

\begin{figure}[t]
    \centering
    \includegraphics[width=0.95\linewidth]{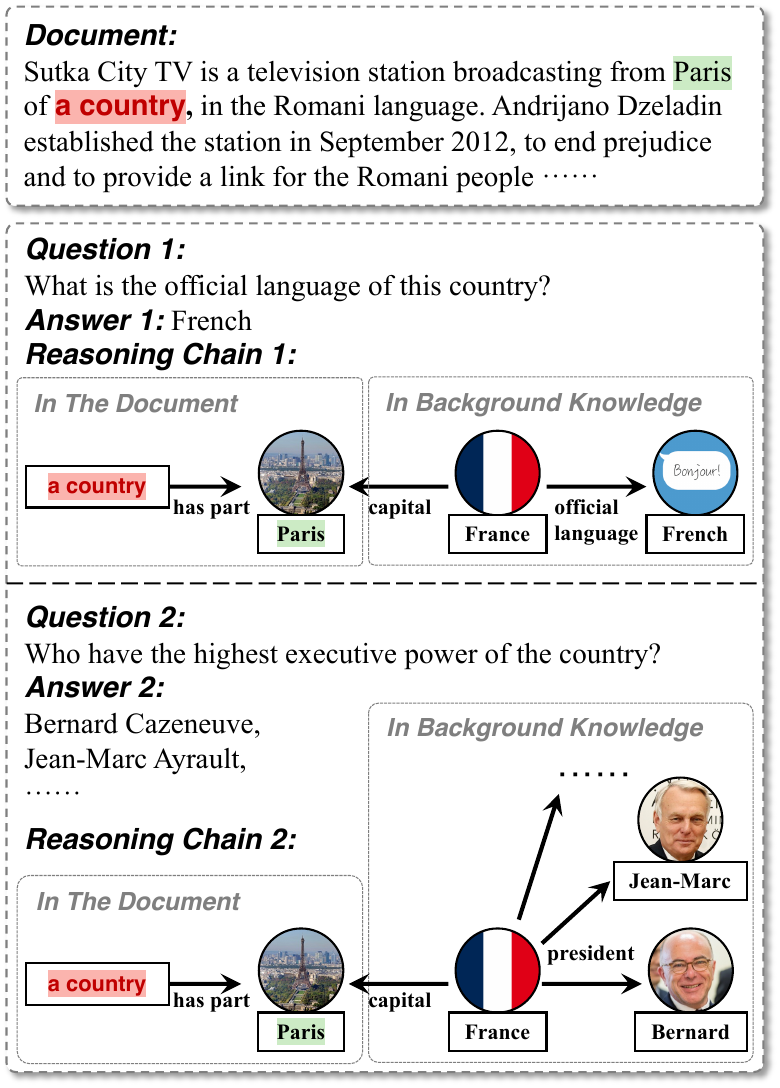}
    \vspace{-0.05in}
    \caption{Examples of \korc. 
    Both question 1 and question 2 require to read the document and make connections to the background knowledge beyond the text.
    % to make connections between the document and the background knowledge base.
    % Readers need to find that the country \textit{has part} Paris.
    % Then, the reader should know that Paris is the capital city of France, and thus reach the answers that are connected to France in the background knowledge.
    }
    \label{fig:intro}
\end{figure}

Deep text understanding requires the integration of text information with its relevant background (prior) knowledge~\citep{gough1986decoding,castles2018ending,smith2021role}.
It has been a long-pursued goal in natural language understanding~\citep{mccarthy1976example,norvig1987unified,huang2019cosmos} for decades, and plays a key role in many real-world applications.

% In viewing of its great significance in real-world applications, many datasets are proposed that play a key role in guiding the development of deep text understanding skills.

Many benchmarks have been proposed to guide the development of deep text understanding skills.
Early attempts formalize text understanding into machine reading comprehension (MRC) framework, such as SQuAD~\cite{rajpurkar2016squad} and RACE~\cite{lai2017race}.
Readers are required to answer questions about the given document in MRC tasks.
Recently proposed benchmarks further highlight the requirement of \textit{deep} text understanding.
To answer their questions, 
% usually requires extra knowledge in addition to the document, 
benchmarks such as CosmosQA~\cite{huang2019cosmos}, DREAM~\cite{sun2019dream}, and C$^3$~\cite{sun2020c3} have tapped into knowledge beyond the text.
Moreover, it is necessary for deep text understanding to reason over a combination of different knowledge sources, as required by QAMP{\small AR}I~\cite{samuel2022qampari} and WikiHop~\cite{welbl2018wikihop}, \textit{etc.}
However, these benchmarks have encountered two limitations.

\vspace{-0.1in}
\paragraph{Limited Knowledge Coverage.}
Many of existing benchmarks are constructed based on knowledge provided by expert annotators (e.g., Q{\small UA}RT{\small Z}~\cite{tafjord2019quartz}) and knowledgeable questions written by question annotators from scratch (\textit{e.g.,} CosmosQA~\cite{huang2019cosmos}).
The discrepancy between the limited background knowledge they cover and massive open-domain knowledge makes it difficult to measure deep text understanding skills at large.
Fortunately, this can be mitigated by generating questions based on large-scale knowledge resources scattered across real-world knowledge bases.

\vspace{-0.1in}
\paragraph{Narrow Answer Space.}
As a compromise for easy construction and evaluation, a large portion of benchmarks ask multiple-choice questions~\cite{lai2017race,sun2019dream} or 
have answers being spans in the provided reading material~\cite{hewlett2016wikireading,welbl2018wikihop,samuel2022qampari}.
% or constitute of multiple-choice questions~\cite{lai2017race,sun2019dream}.
However, multiple-choice questions are processed simply as classification tasks.
Questions based on span-extraction also increasingly become insufficient to challenge the state-of-the-art (SOTA) language models that already show great performance at information extraction~\cite{unifyskg}.

Inspired by the common grounds on deep text understanding, we build a new challenging benchmark, \korc, for \textbf{K}nowledge \textbf{o}riented \textbf{R}eading \textbf{C}omprehension, as shown in Figure~\ref{fig:intro}.
Its most important feature is that both the reading material and external background knowledge are indispensable for every question within \korc.
Readers must connect the document with their equipped prior knowledge and reason across both the text and the background knowledge to reach the final answers.
% be equipped with abundant prior knowledge connecting to the reading material and able to reason over both the text and the background knowledge.

% In viewing of these problems, we build a new challenging benckmark dataset, \korc, for \textbf{K}nowledge \textbf{o}riented \textbf{R}eading \textbf{C}omprehension to further deep text understanding.
% \korc inherits the common grounds on how to develop deep text understanding skills.
% To answer questions in \korc, both the reading materials and external background knowledge are indispensable.
% Readers must be equipped with abundant prior knowledge connecting to the reading material and able to reason over both the text and the background knowledge.
% % readers are required to reason over both the reading materials and external background knowledge.
% Figure~\ref{fig:intro} illustrates examples from \korc.

Different from previous benchmarks, \korc has two advantages.
\textit{\textbf{Broad knowledge coverage}}.
\korc does not require manual knowledge annotation from scratch.
Instead, it uses off-the-shelf knowledge bases as its background knowledge sources
% where we use Wikidata~\cite{denny14wikidata}, a large-scale crowd-sourced knowledge base, 
to guide the construction of knowledgable questions.
% with regard to documents excerpted from Wikipedia.
More exhilaratingly, \korc proves it feasible for LLMs to automatically generate high-quality questions following knowledge instructions.
% verifies the feasibility to automatically generate high-quality questions with large language models (LLMs) with knowledge instructions.
% which means that \korc is potentially extendable without human efforts.
% \korc also adopts a flexible answer space while preserving automatic evaluation metrics.
\textit{\textbf{Flexible answer space}}.
The answers in \korc are labels in knowledge bases, rather than choices or spans from the text.
% Inspired by \citet{welbl2018wikihop} and \citet{samuel2022qampari}, 
In addition, questions in \korc have an in-determinant number of answers (\textit{e.g.,} Question 2 in Figure~\ref{fig:intro}).
We propose two new metrics to facilitate easy evaluation of the variable number of answers.

\korc is constructed based on reasoning chains that weave together documents and background knowledge base.
We provide three versions of \korc based on data annotation methods.
They are \korct from \textbf{T}emplate-based generation, \korch from \textbf{H}uman annotation, and \korcl from \textbf{L}LM annotation.
The final version of \korc contains $9,074$ documents and $31,804$ questions.
% \korc is constructed in three steps.
% First, we prepare documents excerpted from Wikipedia by aligning them to Wikidata via entity linking and document level relation extraction.
% Second, we prepare reasoning chains that weave documents and background knowledge base together and deduce them into question triples.
% Each reasoning chain starts from information mentioned in the documents and ends at entities in the background knowledge base.
% Third, we design three different methods to convert question triples into natural language questions.
% They are \korct from \textbf{T}emplate-based generation, \korch from \textbf{H}uman annotation, and \korcl from \textbf{L}LM annotation.
% The final version of \korc contains $9,256$ documents and $32,390$ questions.
% The test set is splitted into in-domain (IID) and out-of-domain (OOD) subsets according to whether the question triples appear in the training set.
% For fair and automatic evaluation, \korc comes with two newly proposed metrics, which are penalized accuracy (P-ACC) and penalized F1 measure (P-F1).
% \korc also provided detailed evaluation by splitting the test set into in-domain (IID) and out-of-domain (OOD) subsets according to question triples whether  appeared in the training set.
% Dataset statistics.
% Template constructed, human annotated, LLM annotated.
% How to split? 
% How to test (IID and OOD). Evaluation Metric.
We establish the initial baselines for \korc.
We find that even the strongest baseline model only achieves $68.3\% / 30.0\%$ P-F1 (ID / OOD) on \korch, indicating that \korc brings new challenge to natural language understanding.
We also find that LLM-annotated questions in \korcl provide moderate supervision to answer human-generated questions in \korch, which suggests that models can be appropriately instructed to train themselves.
The \korc dataset and codes for our baseline models will be released upon acceptance.

% We establish $4$ kinds of initial baselines for \korc.
% They are fine-tuned language models, chain of thought prompted LLMs, retrieval augmented models, and models jointly reasoning over text and KB.
% Experiment results show that even the most powerful baseline model only achieves {\color{red}$60.5\% / 24.3\%$} P-F1 (IID / OOD) on \korch.
% The large performance discrepancy between IID setting and OOD setting showing that these state-of-the-art models are still \textit{remembering} the answers rather than \textit{learning} to reason with background knowledge.
% We also find that LLM annotated questions provide moderate supervision to answer human generated questions, which suggests that models can be instructed to train themselves.
% The \korc dataset and codes for the baseline models will be released upon acceptance.

% We test \korc upon multiple baselines, including state-of-the-art LLMs. 
% We find that (1) xxx; (2) yyy; (3) zzz.
% Experiment analysis.
% Baseline performance.
% LLM performance.
% IID / OOD performance

% Contributions.
% The dataset / The Dataset Construction Method / The analysis results.
\begin{figure*}[t]
    \centering
    \includegraphics[width=0.98\linewidth]{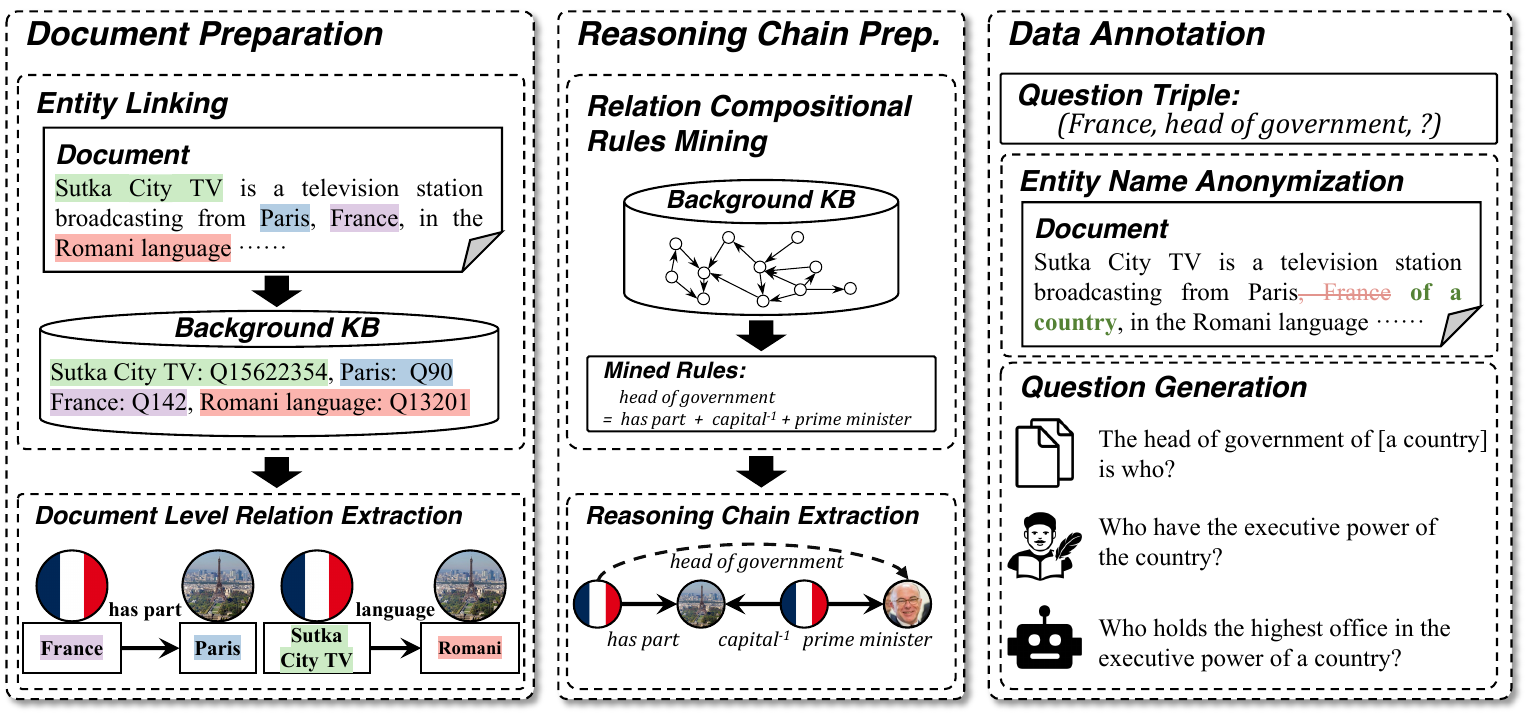}
    % \vspace{-0.1in}
    \caption{
    The overall data collection process.
    In the data annotation step, we also show three real annotation cases from template-based generation, human annotation, and LLM annotation.
    }
    \label{fig:collection}
\end{figure*}

\section{Task Definition}

\korc shares a similar task format with traditional machine reading comprehension (MRC).
The input includes a document $d$ and a natural language question $q$.
Models are required to output the answer $\rva$ to the question after reading the document.

% The key differences between \korc and traditional MRC are three folds.
Different from traditional MRC tasks, \korc presents two key highlights.
Firstly, \korc is augmented with an extra background knowledge base (KB), denoted as $\gK$.
Each semantic triple in the background KB $(e_h, r, e_t) \in \gK$ describes the relation $r$ between the head entity $e_h$ and tail entity $e_t$.
The questions cannot be answered solely within the document or the background KB, but a combination of the two.
% Both the document and the background KB are indispensable.
% (2) The question is about the question entity $e_q$, which is explicitly mentioned in the documents $e_q \in d$ and is anonymized to avoid answer shortcut.
Readers need to reconstruct the reasoning chains, which weaves the document and the background KB together, to find the answers.
Secondly, answers are an in-determinant number of entities in the background KB, \textit{i.e.,} $\rva=\left\{e_i|e_i\in\gK\right\}$, $|\rva| \ge 1$.
Models are encouraged to output neither excessive nor insufficient predictions.
% Readers need to make multi-hop reasoning, which starts from $e_q$ and ends at entities in the background KB.

% We define knowledge oriented reading comprehension dataset as $\gD = \left(\gK, \left\{(d, e_q, \rvq, \rva)\right\}\right)$, where $\gK$ is the background knowledge base (BKB) consisting of semantic triplets $\gK = (e_h, r, e_t)$.

% For a data instance, $d$ is a document containing rich information in text format.
% $e_q$ is the question entity mentioned in the document.
% % and does not appear in the BKB, \textit{i.e.,} $e_q\notin\gK$.
% $\rvq$ and $\rva$ are the question list and the corresponding answer list. 
% Readers are required to answer questions $q_i\in\rvq$ about the question entity $e_q$ mentioned in the document and output the answer $a_i\in\rva$, where $a_i=\left\{e_j|e_j\in\gK\right\}, |a_i| \ge 1$ is a set of one or multiple entities in the BKB.

% In particular, xxxx examples xxx.

\section{Dataset Construction}

\korc requires joint reasoning over text and background KB.
It is constructed in three steps:
(1) We prepare documents and align them to the background KB via entity linking and document level relation extraction;
(2) We prepare reasoning chains that weave documents and background KB together.
We first mine massive relation compositional rules from the background KB and then extract reasoning chains accordingly.
% They start from a question entity $e_q$ mentioned in the document and end at another entity in the background KB.
% They connect information in the document to the background KB;
% based on which questions are asked. 
% The reasoning chain should be contained in both the document and the background KB;
(3) We annotate data by anonymizing the question entity $e_q$ in the document to prevent reasoning shortcut and generate questions based on the reasoning chains.
We design three different methods to annotate the data---template-based generation, human annotation, and large language model annotation.
Figure~\ref{fig:collection} demonstrates the overall data construction process.

\subsection{Step 1: Document Preparation}

% In the document preparation stage, we prepare documents in text format and align the documents to a large-scale knowledge bases containing rich background knowledge.
To provide broad knowledge coverage and facilitate knowledge reasoning, we sample documents from Wikipedia as the reading material and use Wikidata5M~\cite{wang2021wikidata5m}, a subset of Wikidata~\cite{denny14wikidata} consisting of all the entities in Wikipedia, as the background KB.
To align documents from Wikipedia to Wikidata, we need to identify entity mentions in the documents and link them to their entity ID in Wikidata5M (\textit{i.e.,} entity linking).
We also need to extract semantic triples from the documents, which are weaved into the reasoning chains in Step 2.
% are used to construct reasoning chains in the next step.

Fortunately, DocRED~\cite{yao2019docred} provides a large batch of documents from Wikipedia with extracted semantic triples.
Specifically, each document in DocRED is released with extracted entity mentions and relations among the mentions, which comprise semantic triples.
% Entity mentions and their relations constitute semantic triples.
These semantic triples are manually annotated, which have a higher quality than algorithms-extracted ones.
For entity linking, we first link mentions to Wikipedia entities via the existing hyperlink, or use the entity linking toolkit pre-trained on Wikipedia---BLINK~\cite{wu2020scalable}.
Then we use XLORE~\cite{hailong2019xlore} to link Wikipedia entities to Wikidata entities.
In total, $3,291$ documents with valid entity linking results in the training set and validation set of DocRED are used under the grant of MIT License.

\subsection{Step 2: Reasoning Chain Preparation}

A reasoning chain is a list of entities connected by their relations, denoted as $(e_q, r_1, e_1, \cdots, r_n, e_n)$.
In particular, the reasoning chain starts from the document and ends at the background KB, which means $e_q \in d, e_n \in \gK$.
The reasoning chain deduces into a question triple $(e_q, r, ?)$ according to the compositionality of the relations, \textit{i.e.,} $r = r_1 + \cdots + r_n$.
The question triple can be paraphrased into natural language questions like \textit{``Which entities have relation} $r$ \textit{with the question entity} $e_q$\textit{?''}, such that $e_n$ serves as the answer.
% We use Wikidata5M~\cite{wang2021wikidata5m}, a subset of Wikidata consisting of all the entities in the complete English Wikipedia, as the background KB.
To this end, we (1) mine relation compositional rules from massive semantic triples, and then (2) extract reasoning chains from the documents and the background KB according to the compositional rules.

\textbf{Relation Compositional Rule Mining.}
% Compositional rule of relations depicts how multiple relations imply a single relation.
% According to compositional rule, 
% For example, from
Compositional rules of relations are induced from large-scale semantic triples in the background KB.
We use BIMR~\cite{lv2021bimr}, which provides high-quality compositional rules from human annotation.
% with human annotated confidence above $0.3 / 1$.
We supplement more rules mined by AnyBURL~\cite{christian2019anyburl} from the background KB to further increase knowledge coverage.

% Following BIMR, we also supplement rules from AnyBURL~\cite{christian2019anyburl} to automatically mine compositional rules from Wikidata5M and only keep rules with score above $0.01 / 1$.

\textbf{Reasoning Chain Extraction.}
For semantic triple $(e_q, r_1, e_1)$ extracted from document, if a compositional rule $r = r_1 + \cdots + r_n$ exists, we construct the reasoning chain $(e_q, r_1, e_1, \cdots, r_n, e_n)$ and its corresponding question triple $(e_q, r, ?)$.
The resulting reasoning chain satisfies that $e_q$ and $e_1$ are mentioned in the document, \textit{i.e.,} $e_q, e_1 \in d$, and $e_i$ are entities in the background KB, \textit{i.e.,} $e_i \in \gK, i \ge 1$.
% All the constructed reasoning chains are across document and background KB.
$e_1$ serves as the bridge entity between the document and the background KB.

It is worth noting that we filter out reasoning chains which end at the document, \textit{i.e.,} $e_n\in d$, to prevent the reasoning process bypassing the background KB.
The end entity $e_n$ is identified from the document via entity linking.

% Wikidata5M is a subset of  consisting of all the entities in the complete English Wikipedia, which is naturally aligned with the documents in Wikipedia

%  1. Introduce how to construct complex reasoning chain: Rule mining / BIMR + split the reasoning chain into the text and Wikidata5M

\subsection{Step 3: Data Annotation}

Data annotation aims to (1) anonymize the question entity $e_q$ mentioned in the document to prevent reasoning shortcut and (2) generate questions about the anonymized question entity.

In question entity name anonymization, reasoning shortcut means that the document is bypassed and questions can be answered without reading the document.
For example, the answer of questions like \textit{What is the official language of France?} does not require the document as in Figure~\ref{fig:intro}.
Thus, we substitute the mentions of $e_q$ in the document with their anonymized name and polish the document to fluency.
Question name anonymization requires \textit{\textbf{anonymity}} and \textit{\textbf{uniqueness}}.
Anonymity prunes reasoning shortcut and avoids answer leakage.
Uniqueness guarantees that the anonymized name does not refer to other entities mentioned in the text.
% Anonymity prunes the reasoning shortcut.
% Uniqueness means that the anonymized name will not refer to other entities mentioned in the text.
% For example, {\color{red} example}.

The question generation process requires \textit{\textbf{consistency}} and \textit{\textbf{diversity}}.
Semantic information of the natural language question should be consistent with its corresponding question triple.
Besides, diverse syntactic structures for the same relation in different question triples are desired.
For example, question triples $(e_q, r, ?)$, where $r$=\textit{``birth place''} can be converted into \textit{``Where was} $e_q$ \textit{born?''} and \textit{``In which place did} $e_q$ \textit{see the first sunrise of his life?''}.
These two questions expect similar answers though differ in syntactic.

We design $3$ different methods to accomplish the data annotation following the above principles. 

\vspace{-0.1in}
\paragraph{Template-based Generation.}
For question entity anonymization, we substitute entity mentions with their most fine-grained class name in Wikidata.
We also add a unique suffix to the class name to guarantee uniqueness so that it will not refer to entities in the document of the same class.
For question generation, we manually annotate $1-4$ question templates for each relation, which has a placeholder for the question entity.
Given a question triple $(e_q, r, ?)$, the questions are generated via substituting the placeholder in the template of relation $r$ with the anonymized entity name for $e_q$.
We provide example templates in Appendix~\ref{app:template}.

% anonymto guaranteizes the entity mentionss name s by their most fine-grained class name in Wikidata and generate questions according to question templates associated with each relation.
% Class name is a general reference towards multiple different entities, which ensures anonymity.
% To provide uniqueness, we add a unique suffix to the class name to distinguish the question entity from other entities sharing the same class in the documents.
% We manually annotate {\color{red}xx} question templates for each relation, which has a placeholder for the question entity.
% The questions are generated via substituting the placeholder in the template with the anonymized entity name.
% {\color{red}Appendix} gives an example.

% \lipsum[5]

\vspace{-0.1in}
\paragraph{Human Annotation.}
We recruit annotators, who has at least passed Test for English Majors-Band 4 (TEM-4) to annotate the data.
We train them to make sure they are aware of the aforementioned data annotation principles.
% In question entity anonymization, we recommend class names and their synonyms as the optional anonymization name, and ask annotators to select or generate the optimal name.
% In question generation, we encourage the annotators to craft diverse questions according to the semantic information of the question triples.
% The overall process is aided by a visualized annotation platform.
We implement a visualized annotation platform to assist the data annotation process, as shown in Appendix~\ref{app:platform}.

% and generate questions.
% In 

% provides higher quality data than template-based generated data.

% Specifically, for each document, we provide the annotator with the question entity $e_q$ and its associated question triples.
% We recommend several optional anonymization names, which are the class names of the question entity and their synonyms in Wikidata for the annotators and ask them to paraphrase the document after anonymization for fluency.
% We encourage the annotators to craft diverse questions according to the question triples.
% The overall process is aided by a visualized annotation platform.
% {\color{red} Appendix introduce the platform}
% \lipsum[6-7]

\vspace{-0.1in}
\paragraph{Large Language Model Annotation }
is inspired by the success of LLMs in generating datasets~\cite{alisa2022wanli}.
We prompt LLM with demonstrations~\cite{liu2022makes,gpt3} and instructions~\cite{victor2022multitask,jason2022finetuned} to anonymize the question entity, generate questions, and conduct quality inspection.
The provided demonstrations include $2$ manually annotated examples for anonymization and questions.
% In the instruction, we encourage LLMs to generate multiple questions for the same question triple.
% In the quality inspection stage, we input all the generated questions to the LLM and instruct it to select the optimal questions.
In particular, we implement the LLM with \verb|text-davinci-003|, a variant of GPT-3~\cite{gpt3}.
Prompts are shown in Appendix~\ref{app:qgprompt}.

% \lipsum[8]

% \noindent 2. Anonymization of questioning entities with class name

% \noindent 3. Template based question generation

% \noindent 1. Annotation task definition: (1) Anonymization and paraphase; (2) Question generation

% \noindent 2. Annotation platform description (visualization).

After dataset construction, we obtain a total of $9,086$ documents after anonymization and $31,804$ questions.
Notice that each document could have more than one question entities.
They are thus paraphrased into multiple different documents after anonymization.
According to the data annotation method, we present three versions of \korc, namely \korct (Template-based generation), \korch (Human annotation), and \korcl (LLM generation).
We consider \korch as the standard subset of \korc.

\section{Dataset Analysis}

We perform a detailed analysis of \korc.
We first design two evaluation metrics where the number of answers are in-determinant.
Then, we investigate sophisticated data splitting strategy.
Finally, we conduct comprehensive analysis with regard to the data distribution in \korc.

\begin{table*}[tbp]
\centering
\scalebox{0.86}{
\setlength{\tabcolsep}{6pt}
\begin{tabular}{lcccccccccccccc}
\toprule
Split & Train & Valid & Test-ID & Test-OOD & All\\
\midrule
\#Document (Unique) & $7,260$ $(2,332)$ & $4,637$ $(2,074)$ & $546$ $(546)$ & $516$ $(516)$ & $9,086$ $(3,291)$ \\
\#Relation (Unique) & $208$ $(117)$ & $185$ $(113)$ & $121$ $(90)$ & $162$ $(111)$ & $212$ $(119)$ \\
\#Question & $18,945$ & $7,574$ & $3,432$ & $1,853$ & $31,804$ \\
% Average Answer per Question & 3.6 & 3.6 & 4.8 & 2.6 & 3.7 \\
Average Hops per Answer & $2.80$ & $2.80$ & $2.84$ & $2.81$ & $2.80$ \\
\bottomrule
\end{tabular}
}
\caption{Statistics of the final version of KoRC.
Unique documents is the number of documents before anonymization.
Unique relation considers the inverse relation the same as the forward relation.
They are shown in the parenthesis.
}
\label{tab:data}
\end{table*}
\begin{figure*}[htbp]
    \centering
    \includegraphics[width=0.98\linewidth]{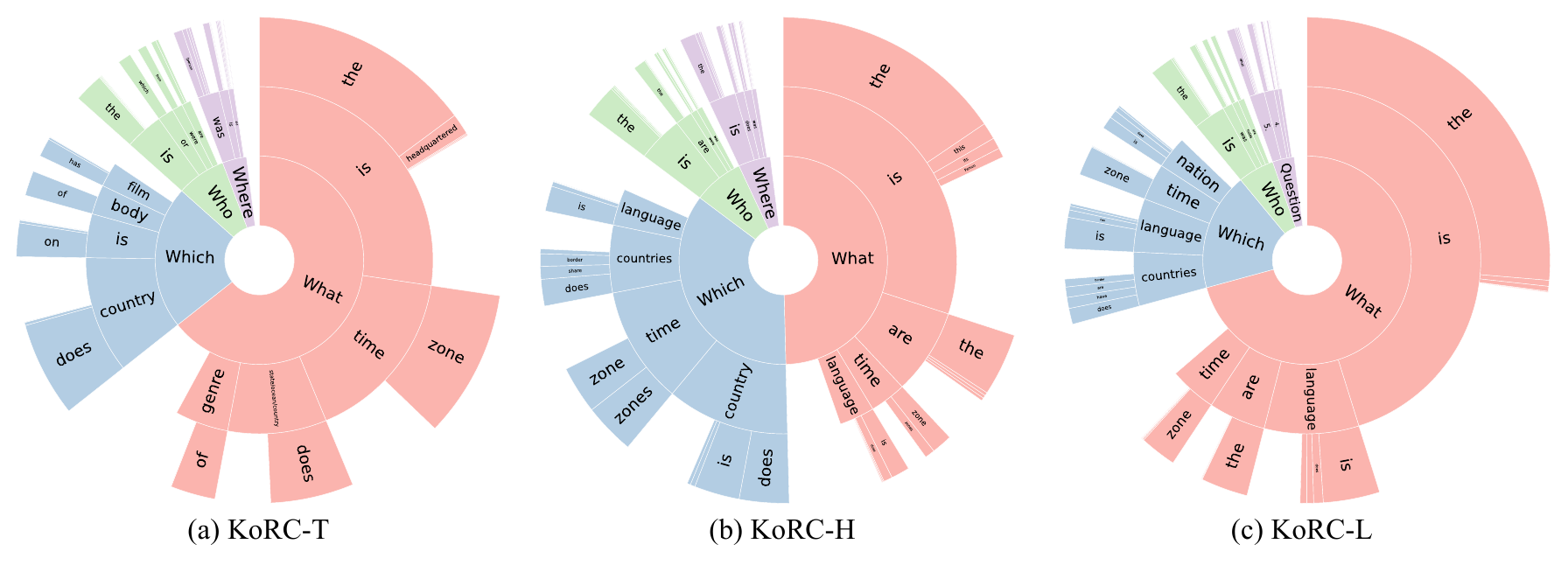}
    \vspace{-0.1in}
    \caption{Distribution of trigram prefixes of questions in \korct, \korch, and \korcl.}
    \label{fig:trigram}
\end{figure*}

\subsection{Evaluation Metric}

We extend exact match accuracy and f1 measure to evaluate machine reading comprehension performance from~\citet{rajpurkar2016squad} by introducing
% \korc is evaluated with 
penalized exact match accuracy (P-ACC) and penalized f1 measure (P-F1).
Since the answer is a set of entities, the metrics need to match the predictions to the ground truth answers with Hungarian algorithm using editing distance.
We define a penalty term in case that the model outputs excessive or insufficient predictions:
$$
\text{penalty} = \frac{\min\{\text{\#prediction}, \text{\#label}\}}{\max\{\text{\#prediction}, \text{\#label}\}}
$$
P-ACC and P-F1 are defined by multiplying the penalty term with the mean accuracy and F1 measure of each matched predictions, respectively.

\subsection{Data Split}
We are mainly concerned with three issues in splitting the data. 
(1) The training set should be sufficient to train a modern MRC model until convergence; 
(2) The test set should avoid any possible data leakage;
(3) How to split the test set into in-distribution (ID) subset and out-of-distribution (OOD) subset for more detailed evaluation?
% Since questions are crafted according to question triples, \korc should leverage the distribution of question triples and provide test in both ID (independent and identically distributed) and OOD (out of distribution) setting.

\begin{figure}[htbp]
    \centering
    \includegraphics[width=0.94\linewidth]{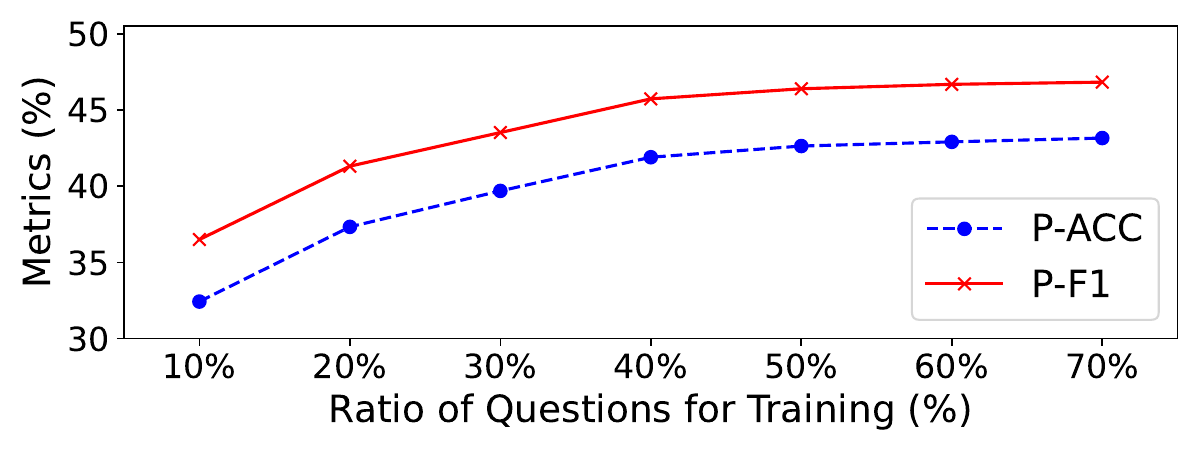}
    \vspace{-0.12in}
    \caption{Training curve.}
    \label{fig:curve}
\end{figure}

\textbf{Training Data Sufficiency.}
We conduct pilot experiment on \korch with BART-base.
We vary the ratio of questions from $10\%$ to $70\%$ for training and use $30\%$ of held-out questions for both validating and testing.
The performance curve is shown in Figure~\ref{fig:curve}, which flattens after $50\%$.
Thus, we use $50\%$ for training.

% We conduct pilot experiments to analyze how much data is required.
% The baseline model we constructed here is based on BART-base, which is a sequence-to-sequence language model. 
% It reads document and question as input, and output the answer entities separated by coma. 
% We held out {\color{red}xxx} documents from \korct for testing, and vary the number of documents for training from {\color{red}xxx} to {\color{red}xxx}.
% The resulting p-acc curve is shown in Figure~\ref{}.
% It shows that the curve become flat when the ratio of documents in the training data arrives at {\color{red}xxx}.
% Thus, we use {\color{red}xxx} documents for training.

\textbf{Leakage Avoidance.}
In the test set, for documents that have multiple question entities, we randomly select one question entity and keep it along with its questions.
The remaining question entities are discarded with their associated questions.
This strategy avoids possible leakage of the name of the anonymized entities.

\textbf{Test Set Splitting.}
Questions in the test set are labeled as ID (OOD) when its question triple $(e_q, r, ?)$ does (not) appear in the training set.
% ID question means its question triple $(e_q, r, ?)$ has appeared in the training set.
% The rest of the questions are labeled as OOD questions.
OOD questions are more challenging than ID questions.

\subsection{Statistic Analysis}

The general statistics of \korc is shown in Table~\ref{tab:data}.
Answers require reasoning chains of an average of $2.80$ hops to reach the answer beyond the document, including the chains within the document.
Figure~\ref{fig:trigram} compares the prefix trigram pattern among different ways of data annotation in Step 3.
It shows that human annotated questions provides the best diversity compared to template based questions and LLM generated questions.
Although LLM annotated questions show lower diversity than template generated questions, we find that LLM can occasional spark novel questions, as the examples shown in Figure~\ref{fig:collection}.

% However, LLM annotated questions still preserve higher diversity than template generated questions, suggesting that LLM annotation is an effective new way to perform automatically question generation.

% Question1: How many passages are required to train a good QA model?

% Question2: Which kind of questions are easier to fuzzy simple baseline methods. (Case study and case statistics / distribution, error classification)

% Question3: DeAnoymization.

% Question4: 
\section{Experiments}

We establish the initial baselines for KoRC and use KoRC to analyze the deep text understanding ability of these baseline models.
More experiments, analysis, and benchmark results are included in the project \href{https://github.com/THU-KEG/KoRC}{repository}.

\subsection{Baseline Models}

We design and implement the initial baselines in the following $4$ categories.

\textbf{Fine-tuned Language Models.}
It has been shown that pre-trained language models are rich in knowledge~\cite{petroni2019language,badr2022lmaskbs}.
Fine-tuning on dataset that requires knowledge reasoning~\cite{alon2020leapofthought,west2022symbolic} elicit the knowledge within LMs.
We view \korc as a sequence-to-sequence task, which can be directly processed by an encoder-decoder language model, such as \textbf{BART-base}~\cite{lewis2020bart} and \textbf{Flan-T5-base}~\cite{hyung2022flant5}.
We also train and evaluate \textbf{Flan-T5-XXL}~\cite{hyung2022flant5}, which scales up to 11B parameters and is trained with task descriptions.
Particularly, the input of the encoder is a concatenation of the anonymized document and the question.
The answers are output as coma separated entity labels.

\textbf{In-Context Learning (ICL) Prompting.}
Prompting is another thread of attempts that stimulate the pre-trained language models to perform complex reasoning task without tuning.
To construct prompts, we use examples in the training set as demonstrations.
% and treat the reasoning chain in the training set as chain of thought.
The demonstration examples are dynamically selected according to sentence similarity of the question and its associated document, which is computed with sentence embedding model MPNet~\cite{kaitao2020mpnet}.
We implement in-context learning prompting with \textbf{GPT-3}~\cite{gpt3} (\verb|text-davinci-002|) and \textbf{GLM-130B}~\cite{aohan2022glm130b}.

\textbf{Retrieval Augmented Models.}
There are opinions on language models alone being insufficient to answer knowledge intensive questions.
To facilitate reasoning requiring knowledge beyond the input text, they propose to augment language models with an external retrieval module, which searchs for the background knowledge from the open-domain Internet, such as RAG~\cite{patrick2020rag}.
We test on \textbf{RAG-seq}, which generates intermediate answers with multiple searching results and synthesis them into the final answer, and \textbf{RAG-token}, which synthesis the searching results and generate the answer.
In \korc, we use the document and the question to search for knowledge and mingle the original document with the searching results to generate the answer.

\textbf{Joint Reasoning over Text and KB.}
These methods align document and questions to the background KB (\textit{i.e.,} Wikidata5M) and perform the knowledge reasoning on the background KB.
\textbf{EmbedKGQA}~\cite{saxena2020improving} converts documents and questions into vectors in the embedding space of the background KB and performs the knowledge reasoning with operations on the embedding vector, where we use ComplEx~\cite{2016complEx}.
We also implement EmbedKGQA with trainable knowledge representations (\textbf{EmbedKGQA$^*$}).
However, limited by computational memory, we only use a subset of the background KB with entities recalled by entity linking.
\textbf{TransferNet}~\cite{shi2021transfernet} uses documents and questions as attention queries in GAT~\cite{velickovic2018graph} to perform explicit knowledge reasoning on the background KB.

\subsection{Main Results}
\begin{table}[!th]
\centering
\scalebox{0.84}{
\begin{threeparttable}
\setlength{\tabcolsep}{3.2pt}
\begin{tabular}{lccccccc}
% \toprule
% \multirow{2}{*}{\textbf{\korct}} & \multicolumn{3}{c}{P-ACC} & \multicolumn{3}{c}{P-F1} \\
% \cmidrule(r){2-4}\cmidrule(lr){5-7}
% & ID & OOD & Mean & ID & OOD & Mean \\
% \midrule
% BART-base  \\
% Flan-T5-base \\
% \midrule
% GPT-3 \\
% GLM-130B \\
% \midrule
% RAG-seq 
% & $60.6$ & $\mathbf{26.7}$ & $48.7$ 
% & $62.1$ & $\mathbf{31.2}$ & $51.3$ \\
% RAG-token 
% & $64.0$ & $24.2$ & $50.0$
% & $65.9$ & $28.4$ & $52.7$ \\
% \midrule
% EmbedKGQA 
% & $\mathbf{66.7}$ & $22.9$ & $\mathbf{51.3}$
% & $\mathbf{73.7}$ & $30.2$ & $\mathbf{58.5}$ \\
% EmbedKGQA$^*$
% & $39.9$ & $15.5$ & $31.3$
% & $46.8$ & $23.4$ & $38.6$ \\
% TransferNet 
% & 35.8 & 14.9 & 28.5
% & 40.7 & 19.2 & 33.2 \\
% \bottomrule
\toprule
\multirow{2}{*}{\textbf{\korch}} & \multicolumn{3}{c}{P-ACC} & \multicolumn{3}{c}{P-F1} \\
\cmidrule(r){2-4}\cmidrule(lr){5-7}
& ID & OOD & Mean & ID & OOD & Mean \\
\midrule
BART-base
& $50.3$ & $24.9$ & $41.4$
& $52.9$ & $30.2$ & $44.9$ \\
Flan-T5-base 
& $33.5$ & $24.0$ & $30.2$
& $35.8$ & $27.5$ & $32.9$ \\
Flan-T5-XXL 
& $\mathbf{63.8}$ & $\mathbf{32.3}$ & $\mathbf{52.8}$
& $65.8$ & $\mathbf{37.2}$ & $\mathbf{55.8}$ \\
\midrule
GPT-3 
&  $18.2$ &  $24.6$ & $20.5$
&  $22.2$ & $30.2$ & $25.0$ \\
GLM-130B
&  $9.9$ &  $14.9$ & $11.6$
&  $12.7$ & $18.8$ & $14.8$ \\
\midrule
RAG-seq
& $61.7$ & $25.9$ & $49.2$
& $63.7$ & $30.0$ & $51.9$ \\
RAG-token 
& $57.4$ & $23.5$ & $45.5$
& $59.1$ & $27.2$ & $47.9$ \\
\midrule
EmbedKGQA 
& $61.2$ & $21.9$ & $47.4$ 
& $\mathbf{68.3}$ & $28.9$ & $54.5$ \\
EmbedKGQA$^*$ 
& $34.0$ & $13.6$ & $26.9$
& $41.6$ & $21.8$ & $34.6$ \\
TransferNet 
& $32.7$ & $12.9$ & $25.8$
& $37.7$ & $16.6$ & $30.3$ \\
\bottomrule
% \toprule
% \multirow{2}{*}{\textbf{\korcl}} & \multicolumn{3}{c}{P-ACC} & \multicolumn{3}{c}{P-F1} \\
% \cmidrule(r){2-4}\cmidrule(lr){5-7}
% & ID & OOD & Mean & ID & OOD & Mean \\
% \midrule
% BART-base  \\
% Flan-T5-base \\
% \midrule
% GPT-3 \\
% GLM-130B \\
% \midrule
% RAG-seq \\
% RAG-token 
% & $56.8$ & $21.8$ & $44.5$
% & $58.6$ & $25.7$ & $47.1$ \\
% \midrule
% EmbedKGQA 
% & $62.7$ & $22.4$ & $48.6$
% & $69.7$ & $29.2$ & $55.5$ \\
% EmbedKGQA$^*$ 
% & $42.8$ & $18.9$ & $34.4$
% & $49.6$ & $26.0$ & $41.3$ \\
% TransferNet 
% & $31.8$ & $12.7$ & $25.1$
% & $36.8$ & $16.2$ & $29.6$ \\
% \bottomrule
\end{tabular}
\end{threeparttable}
}
\caption{
Baseline results on \korch.
% EmbedKGQA$^*$ means that it updates the knowledge representations during training, while EmbedKGQA uses freezed knowledge representations.
Baseline results on \korcl and \korcl are shown in Appendix~\ref{app:moreresult}.
}
\label{tab:baseline}
\end{table}

Table~\ref{tab:baseline} shows all the baseline results on \korch---the standard subset of \korc.
The strongest baseline achieves $52.8\%$ average P-ACC and $55.8\%$ average P-F1 by Flan-T5-XXL, which suggests that fine-tuned large language models have strong capability to use background knowledge.
RAG-seq and EmbedKGQA also achieve competitive performance,
% The best results are achieved by either RAG-seq or EmbedKGQA in all the evaluation metrics.
which have the ability to retrieve background knowledge from the open-domain Internet or access the background KB.
Although language model pre-training brings large-scale knowledge into the model, ICL prompted LLMs do not provide a satisfactory performance on \korc, which indicates that precise recalling of background knowledge plays a key role in answering our questions.
These results show that \korc serves its designing purpose to test deep text understanding skills.

Evaluation results show a performance drop around $20\%-40\%$ from ID set to OOD set on \korch.
This discrepancy suggests that these models mainly learn to \textit{\textbf{remember}} the answers, rather than \textit{\textbf{generalize}} to different query triples.
Meanwhile, knowledge representation based EmbedKGQA is superior
% shows slight advantage 
or comparable to knowledge retrieving based RAG-seq on ID sets while it is outmatched on OOD sets.
This occurs because knowledge representations are constructed based on relation compositional rules, thus easy to overfit the ID questions.
Splitting the test set in \korc provides a new way to evaluate the true deep text understanding skills.

ICL prompted LLMs are observed to perform better on the OOD set than the ID set.
This counter-intuitive result is caused by the notorious repetition problem~\cite{repetition}.
ID shares a similar distribution to the training set so LLMs directly copy the results from the demonstrations, while the OOD set urges the model to think independently.
Another abnormal model is EmbedKGQA$^*$.
Although its knowledge representation can be updated, it falls short of EmbedKGQA by a large margin due to its limited background knowledge that can be held into the random access memory of GPUs, which further reflects the broad knowledge coverage of \korc.

% We conduct all the experiments evenly across \korct, \korch, and \korcl.
% The detailed experimental results are shown in Table~\ref{tab:morebaseline} attached to the appendix.
% Although these is no xxxx

% Leave a paragraph to analyze LLMs.

\subsection{Cross Evaluation}

% \begin{table}[h]
% \centering
% \scalebox{0.85}{
% \begin{tabular}{llrrrrrrrr}
% \toprule
% % \textbf{BART-base}
% & &  \korct & \korch & \korcl \\
% \midrule
% \multirow{3}{*}{\scalebox{0.9}{\rotatebox{90}{\small \textbf{BART-base}}}}
% & \korct \\
% & \korch \\
% & \korcl \\
% \midrule
% \multirow{3}{*}{\rotatebox{90}{\small \textbf{GPT-3}} }
% & \korct \\
% & \korch \\
% & \korcl \\
% \midrule
% \multirow{3}{*}{\rotatebox{90}{\small \textbf{RAG-seq}} }
% & \korct \\
% & \korch & $46.5$ & $51.9$ & $47.9$ {\color{violet}$(4.0\downarrow)$} \\
% & \korcl \\
% \midrule
% \multirow{3}{*}{\scalebox{0.77}{\rotatebox{90}{\small \textbf{EmbedKGQA}}} }
% & \korct \\
% & \korch \\
% & \korcl \\
% \bottomrule
% \end{tabular}
% }
% \caption{Cross evaluation results in P-F1 (\%).
% The left most column shows where the training data comes from.
% }
% \label{tab:cross}
% \end{table}

\begin{table}[t]
\centering
\scalebox{0.82}{
\setlength{\tabcolsep}{2.8pt}
\begin{tabular}{lccccccc}
\toprule
\textbf{BART-base} &  \korct & \korch & \korcl \\
\cmidrule(lr){1-4}
\korct 
& $\mathbf{48.7}$ & $39.4$ {\color{violet}$(9.3\downarrow)$} & $37.5$ {\color{violet}$(11.2\downarrow)$} \\
\korch 
& $41.7$  {\color{violet}$(3.2\downarrow)$} & $\mathbf{44.9}$ & $40.8$ {\color{violet}$(4.1\downarrow)$} \\
\korcl 
& $40.7$ {\color{violet}$(6.4\downarrow)$} & $42.3$ {\color{violet}$(4.8\downarrow)$} & $\mathbf{47.1}$ \\
\midrule
\textbf{GPT-3} &  \korct & \korch & \korcl \\
\cmidrule(lr){1-4}
\korct
& $\mathbf{24.5}$ & $23.6$ {\color{violet}$(0.9\downarrow)$} & $23.2$ {\color{violet}$(1.3\downarrow)$} \\
\korch 
& $23.7$ {\color{violet}$(1.3\downarrow)$} & $\mathbf{25.0}$ & $24.9$ {\color{violet}$(0.1\downarrow)$} \\
\korcl
& $23.0$ {\color{violet}$(0.9\downarrow)$} & $23.8$ {\color{violet}$(0.1\downarrow)$} & $\mathbf{23.9}$ \\
\midrule
\textbf{RAG-seq} &  \korct & \korch & \korcl \\
\cmidrule(lr){1-4}
\korct
& $\mathbf{51.3}$ & $40.8$ {\color{violet}$(10.5\downarrow)$} & $38.6$ {\color{violet}$(12.7\downarrow)$} \\
\korch 
& $46.5$ {\color{violet}$(5.4\downarrow)$} & $\mathbf{51.9}$ & $47.9$ {\color{violet}$(4.0\downarrow)$} \\
\korcl
& $46.7$ {\color{violet}$(8.2\downarrow)$} & $48.1$ {\color{violet}$(6.8\downarrow)$} & $\mathbf{54.9}$ \\
\midrule
\textbf{EmbedKQGA} &  \korct & \korch & \korcl \\
\cmidrule(lr){1-4}
\korct
& $\mathbf{58.5}$ & $44.1$ {\color{violet}$(14.4\downarrow)$} & $38.5$ {\color{violet}$(20.0\downarrow)$} \\
\korch 
& $53.6$ {\color{violet}$(0.9\downarrow)$} & $\mathbf{54.5}$ & $47.8$ {\color{violet}$(6.7\downarrow)$} \\
\korcl
& $49.5$ {\color{violet}$(6.0\downarrow)$} & $51.5$ {\color{violet}$(4.0\downarrow)$} & $\mathbf{55.5}$ \\
\bottomrule
\end{tabular}
}
\caption{Cross evaluation results among \korct, \korch, and \korcl in terms of P-F1 (\%) averaged over IID set and OOD set.
The left most column shows where the training data are from.
}
\label{tab:cross}
\end{table}

We conduct cross evaluation among \korct, \korch, and \korct to verify whether automatically generated questions can be used as distant supervision to learn deep text understanding skills.
In particular, we train models on one of the three versions of datasets, and evaluate on the test set of all the three versions.
Cross evaluation results are shown in Table~\ref{tab:cross}.

As expected, all the cross evaluation results drop compared to the those where training data and test data are produced by the same data annotation method.
Nevertheless, among all the three versions, \korch brings more sophisticated deep text understanding skills to the model, with even as marginal as a $0.9\%$ performance drop for EmbedKGQA on \korct in terms of average P-F1.
This is attributed to the diversity of the questions generated by our annotators.
Meanwhile, training on \korcl only results in a moderate performance drop on \korct and \korch.
By contrast, models trained on \korct struggle with test questions in \korch and even \korcl.
This suggests a feasibility to instruct LLMs with massive real-world knowledge to generate high-quality questions.
These questions can then be used as distant supervision to train models to achieve deep language understanding.

% which brings insight that we can , which in turn used to train models to achieve deep language understanding.

% shows that all the three datasets have result in 
% human annotated dataset \korch ejects more sophisticated deep text understanding skills into the models.

\subsection{Analysis}

We further conduct empirical analysis on \korc, including error analysis and ablation study.

\begin{figure}[t]
    \centering
    \includegraphics[width=0.98\linewidth]{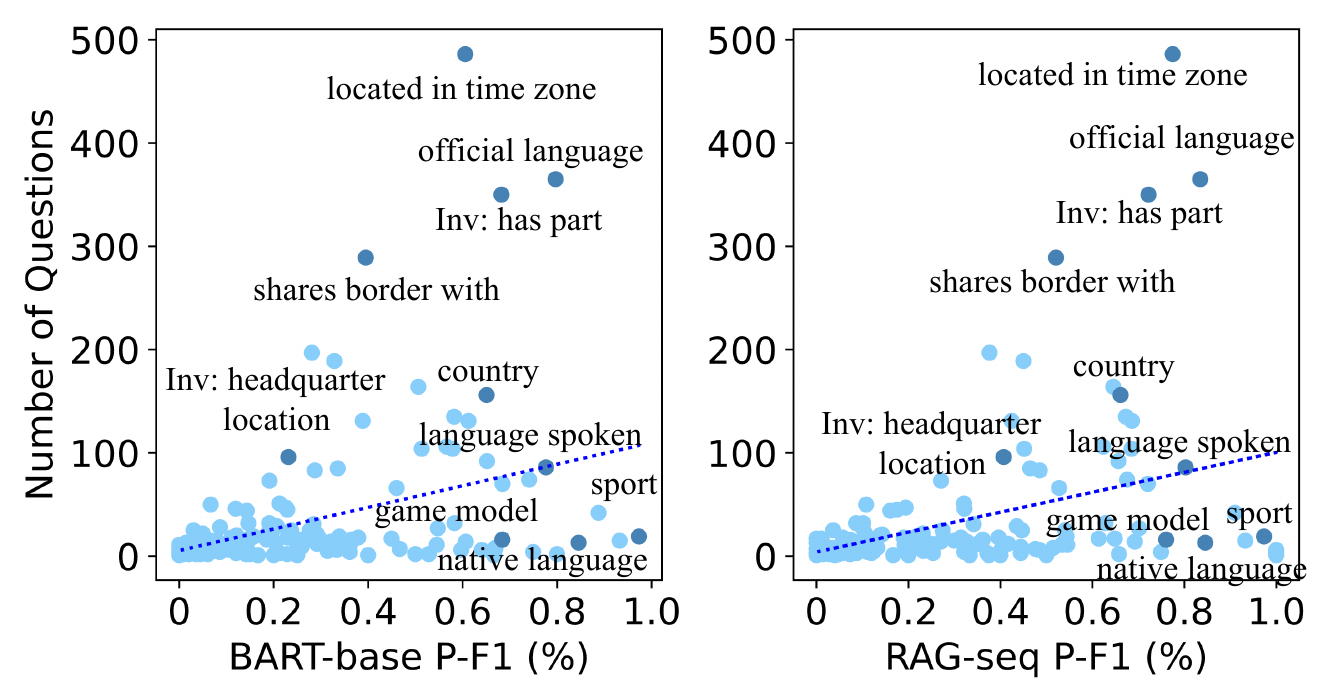}
    % \vspace{-0.1in}
    \caption{Error analysis.
    Each point corresponds to a relation with its number of questions in \korc and average P-F1 recorded on BART-base and RAG-seq.
    The dashed lines indicate linear regression results.
    We highlight and label several representative relations.
    }
    \label{fig:error}
\end{figure}

\textbf{Error Analysis.}
Each question in \korch corresponds to a question triple $(e_q, r, ?)$, which contains a relation $r$.
We examine the error distribution with regard to relations.
Figure~\ref{fig:error} plots the scatter charts for each relation in \korc.
Each point represents a relation with its question number and average P-F1 on BART-base and RAG-seq.

To better demonstrate the correlation between question number and P-F1, we run least square error regression and show in dashed line.
The regression results indicate the trend that relations with fewer questions (long tail relations) are more difficult than relations with abundant questions.
However, there are outlier relations scattered in the top left (bottom right) corner, which means they have many (few) questions in \korc that are difficult (easy) to answer.
We label a few of these outlier relations in Figure~\ref{fig:error}.
We find that top-left-relations are mostly equipped with multiple answers.
For example, questions involving the inverse relation of \textit{headquarter location} usually ask \textit{Which organizations are headquartered in this place?} are difficult to recall all the correct answers.
For the bottom-right relations, they usually construct single-answer questions, such as \textit{native language} and \textit{sport}.

\begin{table}[t]
\centering
\scalebox{0.85}{
\begin{tabular}{ccccccc}
\toprule
\textbf{\korch} & Original & -Document & -Anon.\\
\midrule
\textbf{BART-base}
& $44.9$ & $24.5$ {\color{violet}$(20.4\downarrow)$} & $55.1$ {\color{violet}$(10.2\uparrow)$} \\ 
\bottomrule
\end{tabular}
}
\caption{Ablation results on \korch with BART-base in terms of P-F1 (\%) averaged over IID and OOD sets.}
\label{tab:ablation}
\end{table}

\textbf{Ablation.}
We remove documents from \korch, which makes \korch degenerate into a question answering benchmark.
We also experiment whether the entity name will result in reasoning shortcut without anonymization.
The original name of the question entity is appended to the document.
Table~\ref{tab:ablation} shows the ablation study results.

We find that removing document significantly undermines the results of BART-base with a performance drop at $20.4\%$ in P-F1.
This shows that text information is indispensable in \korc.
Readers are not encouraged to directly answer the questions without reading the given document.
When we provide the entity name as part of the reading material, 
the P-F1 of BART-base increases from $44.9\%$ to $55.1\%$.
This shows that entity name contains direct clues to answering the question and annotating anonymized entity name  cannot be omitted.

\section{Related Work}

% \textbf{Deep Text Understanding.}

% Joint reasoning with text and BGK: baseline. (survey + baseline)
\paragraph{Machine Reading Comprehension.}
Devising intelligent systems to answer
questions on knowledge in text form has long been a challenge in Natural Language Understanding (NLU)~\cite{welbl2018wikihop}, and the MRC task plays an important part in evaluating NLU~\cite{Ho2022ASO}.
Abundant datasets have been proposed to advance research in MRC. One of the earliest work is MCTest~\cite{Richardson2013MCTestAC}, a multiple-choice reading comprehension dataset. 
Following works have surged to advance more challenging text understanding with more complicated answer formats. 
Based on the answer format, MRC datasets can by grouped into four types: span extraction~\cite{hewlett2016wikireading,welbl2018wikihop,samuel2022qampari}, multiple-choice~\cite{sun2019dream,tafjord2019quartz,huang2019cosmos,samuel2022qampari}, cloze style~\cite{mostafazadeh2016rocstory}, and free-form~\cite{khashabi2018multirc} answer. 

\paragraph{Deep Text Understanding.}
Background knowledge integration is regarded as the key ingredient of deep text understanding. 
% Deep text understanding requires not only understanding what is stated explicitly in text, but also reading between the lines, i.e., integrating text information with its relevant background knowledge. 
Different kinds of background knowledge have been employed, such as commonsense knowledge (e.g., ATOMIC~\cite{Sap2019ATOMICAA}), and world knowledge (e.g., Wikidata~\cite{denny14wikidata}). Representative works include WikiReading~\cite{hewlett2016wikireading} which aims to predict textual values from Wikidata by reading the corresponding Wikipedia text, DREAM~\cite{sun2019dream} whose questions requires unspoken commonsense knowledge, Q{\small UA}RT{\small Z}~\cite{tafjord2019quartz} that requires understanding and applying qualitative knowledge, and CosmosQA~\cite{huang2019cosmos} that requires contextual commonsense reasoning.

% In addition, to evaluate the
% multi-step reasoning ability of the models across
% paragraphs, Welbl et al. (2018) introduce the multihop MRC task. It requires a model to answer a
% given question by performing reasoning over multiple paragraphs. Recently, conversational MRC
% tasks such as Choi et al. (2018) and Reddy et al.
% (2019) have also been introduced.

% Two trends to deeper text understanding:
% 1) answer types.
% 2) reasoning path.
%     Multi-hop MRC.
% 3) Background knowledge
    
%     knowledge bases

Compared with the existing datasets, \korc is constructed with the instruction from real-world large-scale knowledge base.
The answers of our \korc are labels in the knowledge bases, and the number of answers is in-determinant, challenging MRC more. 
Most importantly, both the reading materials and external background knowledge are indispensable for every question in \korc, which prevents reasoning shortcut effectively.

\section{Conclusion}

In this paper, we propose a new benchmark---\korc for deep text understanding with broad knowledge coverage and flexible answer format.
Our contributions are not only the dataset itself, but also we demonstrate the feasibility to guide LLMs to generate deep text understanding questions with the help of large-scale background KB.
Our baseline experiments demonstrates to which extent existing powerful models can leverage background knowledge to understand passages by trying to solve \korc.
% show that baseline models are far from satisfactory to solve \korc.
In the future, we plan to extend \korc to more complicated knowledge, such as literal knowledge and qualifier knowledge in common knowledge bases.
It is intriguing to design more skillful reader models via connecting the document with background knowledge.

% The overall contributions of this paper are thus three folds.
% (1) The dataset \korc.
% (2) The dataset construction methods.
% (3) The analysis results for state-of-the-art models based on \korc.

\section*{Limitations}
We propose and construct \korc as a new benchmark dataset for deep text understanding.
The limitations are two folds.
First, in the benchmark design, \korc do not take more complicated knowledge into consideration, including literal knowledge and qualifier knowledge.
We leave extending \korc to these knowledge in future work.
Second, in the dataset construction, we examine automatic name anonymization and question generation strategy, and present \korcl.
\korcl relies on large language models.
Rather than medium-scaled language models that can be maintained by a single machine, GPT-3 is used via its online APIs.
Although the service of GPT-3 is currently available, we still need to find a substitution for better reproducibility.
Besides, although LLM saves human effort, the execution of LLMs potentially consumes more energy power.
It would be better if we can preserve the high question generation quality and propose a small model to proceed data annotation.

\section*{Ethics Statement}
Our proposed dataset, \korc, is constructed with the knowledge guidance from Wikidata.
As a crowd-sourced knowledge base, it is possible that Wikidata contains bias knowledge and even poisonous information.
For example, Wikidata contains more information in the English.
It is possible that \korc also inherit the bias from Wikidata.
Another ethical concern raises from the payment of our annotators.
All the annotators are payed equally according to the number of documents and questions they annotated.
We hope that \korc can be properly used to guide the development of deep text understanding models after we release it.


\bibliography{1-ref}
\bibliographystyle{acl_natbib}

\clearpage

\appendix

\section{Data Annotation Details}
\subsection{Question Templates}\label{app:template}

In Section 3.3, we introduced three different ways to annotate data. 
They are template-based generation, human annotation, and LLM generation.
Here we supplement more technical details on these three methods.

\begin{table*}[t]
\centering
\scalebox{0.88}{
\begin{tabular}{lllllll}
\toprule
Relation Direction & \multicolumn{1}{l}{Relation Label}                 & Template                                      \\ \midrule
\multirow{8}{*}{Forward}&\multirow{2}{*}{$r = $ \textit{member of political   party}} 
& What political party was {[}$e_q${]} a member of? \\
&& Which political party does {[}$e_q${]} belong to? \\
\cmidrule(lr){2-3}
&\multirow{2}{*}{$r = $ \textit{place of burial}}        
& Where is the burial place of {[}$e_q${]}? \\
&& Where was {[}$e_q${]} buried after his/her death?     \\
\cmidrule(lr){2-3}
&\multirow{2}{*}{$r = $ \textit{cast member}}                 
& {[}$e_q${]} is a cast member of which movie?      \\
&& What movies or work has {[}$e_q${]} been in?      \\
\cmidrule(lr){2-3}
&\multirow{2}{*}{$r = $ \textit{country of citizenship}}         & Which country does [x] come from?      \\
&& What nationality does [x] hold?      \\
\cmidrule(lr){1-3}
\multirow{4}{*}{Forward}&\multirow{2}{*}{$r = $ \textit{Inv: producer}}
& Which work is produced by {[}$e_q${]}?      \\
&& Which work did {[}$e_q${]} produce?      \\
\cmidrule(lr){2-3}
&\multirow{2}{*}{$r = $ \textit{Inv: parent organization}}
& Whose parent organization is {[}$e_q${]}?      \\
&& Which subsidiaries does {[}$e_q${]} have?    \\
\bottomrule
\end{tabular}
}
\caption{Example question templates for data annotation of \korct.}
\label{tab:template}
\end{table*}

\subsection{Human Annotation Details}

\subsubsection{Annotator Recruiting}
\label{app:annotator}

We recruit professional annotators who have English as their second language.
These annotators are employees of data provider.
All the annotators working for \korch have passed Test for English Majors-Band 4 (TEM-4).
In particular, TEM-4 is a national Test for students majoring in English in the end of their
second year at university in China.
This qualification ensures that they can correctly read our document, paraphrase the document after anonymization, and write fluent questions according to the question triples.

\subsubsection{Annotation Platform}
\label{app:platform}

We design visualized annotation platform to help annotators to better annotate data.
The annotation platform aims to (1) track editing history and (2) provide knowledge information such as anonymization name recommendations.

\begin{figure*}[t]
    \centering
    \includegraphics[width=0.95\linewidth]{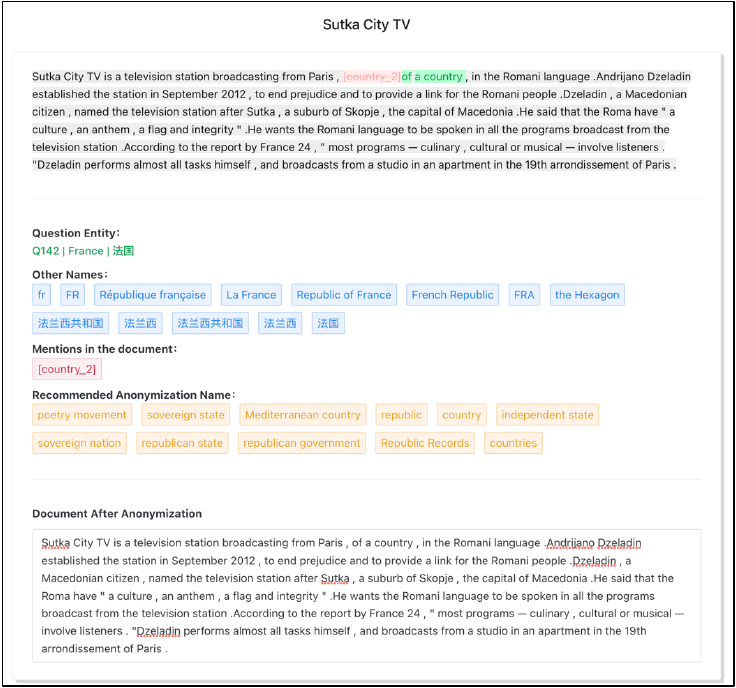}
    \caption{Screenshot of our annotation platform for entity name anonymization.}
    \label{fig:docs}
\end{figure*}

\paragraph{Entity Name Anonymization.}
Figure~\ref{fig:docs} shows the screenshot of our GUI for entity name anonymization.
The annotators are asked to anonymize the question entities by modify the input box right below ``Document After Anonymization''.
We provide information, including question entity names, entity mentions, and recommended anonymization name in colored cards.
Annotators could easily identify which spans are deleted (marked by red background) and which spans are newly added (marked by green background).
In the screenshot, we delete span \colorbox{red!10}{{[}country\_2{]}} and add span \colorbox{green!15}{{of a country}}.

\begin{figure*}[t]
    \centering
    \includegraphics[width=0.95\linewidth]{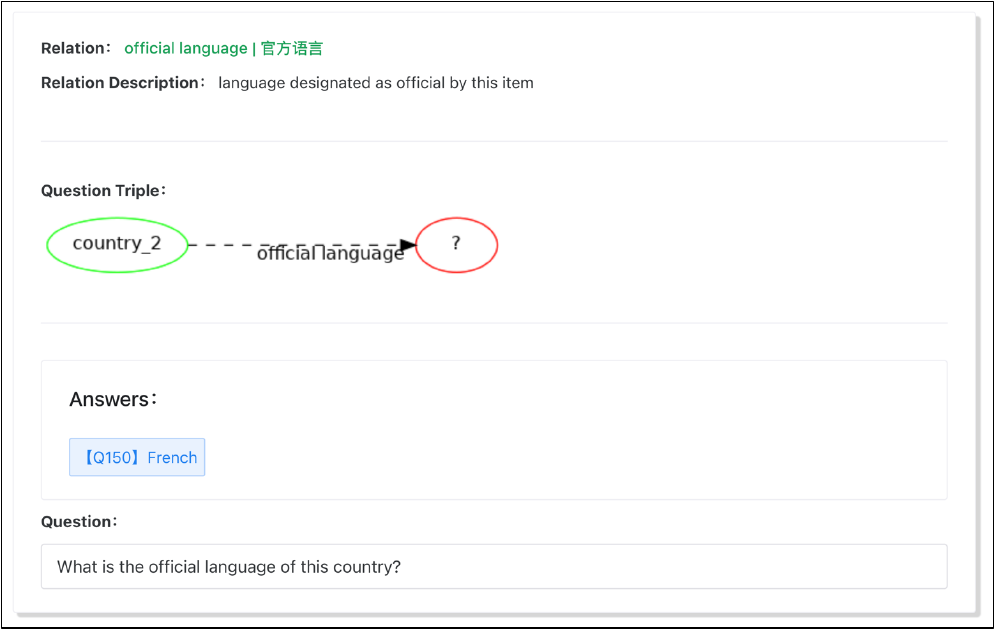}
    \caption{Screenshot of our annotation platform for question generation.}
    \label{fig:question}
\end{figure*}

\paragraph{Question Annotation.}
Figure~\ref{fig:question} shows the screenshot for question annotation.
The annotators are provided with the question triple and the corresponding answers.
They are required to write questions accordingly.

\subsection{Prompt Design for LLM Annotation}
\label{app:qgprompt}

We use in-context learning to instruct LLMs, where we use GPT-3, to proceed data annotation.
For entity name anonymization, we provide LLM with the class name of the question entity and ask LLM to select the optimal class name, which will not leak any information to the answer, to paraphrase the document.
For question generation, we first instruct LLM to generate multiple candidate questions.
Then, we design another instruction to select the optimal questions, which is similar to the quality control step in data engineer.

\begin{table*}[t]
\centering
\scalebox{0.85}{
\begin{tabular}{p{1.0\linewidth}ccccc}
\toprule
Prompt for forward relation. \\
\midrule
\textbf{Instruction:} A semantic triple describe the relation between one head entity and one tail entity. For example, Job Biden -> native language -> English is one semantic triple which means Job Biden (head entity)'s native language (relation) is English (tail entity), now you are given one incomplete semantic triple where the tail entity is missing and one hint which would tell what all the possible missing entity is. your task is to design 5 questions based on the given semantic triple and the hint to find out the missing tail entity. \\
\textbf{Notice}: the given hint could be utilized to design more accurate questions with respect to the given possible missing entities, but any part of the hint should not be contained in the generated question! \\
\\
\textbf{Example 1}: \\
\textbf{Input}: \\
\textbf{Question Triple}: independent state F -> shares border with (countries or administrative subdivisions, of equal level, that this item borders, either by land or water. A single common point is enough.) -> missing entity \\
\textbf{hint}: possible missing entity could be: "Paraguay","Chile","Uruguay","Bolivia","Brazil" \\
 \\
\textbf{Output}: \\
1. Which countries does independent state F border? \\
2. What countries do the boundaries of independent state F touch? \\
3. Who are the neighboring countries of independent state F? \\
4. What states share a border with independent state F? \\
5. To which countries does independent state F have a frontier? \\
\\
\textbf{Example 2}: \\
\textbf{Input}: \\
\textbf{Question Triples}: person F -> occupation (occupation of a person; see also "field of work" \\(Property:P101), "position held" (Property:P39)) -> missing entity \\ 
\textbf{hint}: possible missing entity could be: "actor", "singer" \\
% \textbf{Question Triple}: \colorbox{blue!10}{{[}\textit{Question Triples to Fill}{]}} \\
% \textbf{hint}: \colorbox{blue!10}{{[}\textit{hint to Fill}{]}} \\
\\
\textbf{Output:} \colorbox{red!10}{{[}\textit{LLM output}{]}} \\
\bottomrule
\end{tabular}
}
\caption{Prompt for generating questions involved with forward relation.
}
\label{tab:generateprompt1}
\end{table*}

\begin{table*}[t]
\centering
\scalebox{0.85}{
\begin{tabular}{p{1.0\linewidth}ccccc}
\toprule
Prompt for inverse relation. \\
\midrule
\textbf{Instruction}: A semantic triple describe the relation between one head entity and one tail entity. For example, Job Biden -> native language -> English is one semantic triple which means Job Biden (head entity)’s native language (relation) is English (tail entity), now you are given one incomplete semantic triple where the head entity is missing and one hint which would tell what all the possible missing entity is. your task is to design 5 questions based on the given semantic triple and the hint to find out the missing head entity. \\
\textbf{Notice}: the given hint could be utilized to design more accurate questions with respect to the given possible missing entities, but any part of the hint should not be contained in the generated question! \\
\\
\textbf{Example 1}:\\
\textbf{Input}:\\
\textbf{Question Triple}: missing entity -> has part(s) (part of this subject; the inverse property of "part of" (P361). See also "has parts of the class" (P2670).) -> country A\\
\textbf{hint}: possible missing entity could be: "Northern America", "North American Football Union", "G20", "Allies of the Second World War", "Procurement G6", "North America"\\

\textbf{Output}:\\
1. What international organizations and events have country A participated in?\\
2. What international congregations and activities have the country A partaken in?\\
3. To what foreign associations and interactions have country A contributed?\\
4. What external associations and proceedings have country A been a part of?\\
5. What associations and episodes on the international level have country A been a part of?\\
\\
\textbf{Example 2}: \\
\textbf{Input}: \\
\textbf{Question Triple}: missing entity -> award received (award or recognition received by a person, organisation or creative work) -> order of chivalry \\
\textbf{hint}: possible missing entity could be: "Theobald Bethmann-Hollweg", "Abdul Karim", "Abraham Moyshevich Hekkelman", "Gerald Lloyd-Verney", "Faisal of Saudi Arabia", "Peter Westmacott", "John Simon, 1st Viscount Simon", "Johan E. Mellbye", "Francisco Craveiro Lopes", "Alfred Munnings", "Vyvyan Holt", "Arthur Sullivan", "Mary Curzon, Baroness Curzon of Kedleston" \\
% \textbf{Input}: \\
% \textbf{Question Triple}: \colorbox{blue!10}{{[}\textit{Question Triples to Fill}{]}} \\
% \textbf{hint}: \colorbox{blue!10}{{[}\textit{hint to Fill}{]}} \\
\\
\textbf{Output:} \colorbox{red!10}{{[}\textit{LLM output}{]}} \\
\bottomrule
\end{tabular}
}
\caption{Prompt for generating questions involved with inverse relation.}
\label{tab:generateprompt2}
\end{table*}

\paragraph{Question Generation.}
Prompts for question generation are shown in Table~\ref{tab:generateprompt1} and Table~\ref{tab:generateprompt2}.
Notice that for question triples involving forward relations and inverse relations, we design different prompts.
They are mainly different in the example.

\begin{table*}[t]
\centering
\scalebox{0.9}{
\begin{tabular}{p{0.95\linewidth}ccccc}
\toprule
Prompt for question selection. \\
\midrule
\textbf{Instruction}: You are given several questions, which share similar semantics and same answers. Their corresponding answers are also provided. Your task is to pick out the most accurate, the smoothest, the most novel question from the given questions with respect to given answers based on the given information. Notice, any part of the corresponding answers should not be contained in the selected question and the selected question should not be simply answered by "yes" or "no"! \\
\\
1. What language(s) does the person speak?\\
2. What language(s) can the person read, write and sign?\\
3. What language(s) is the person familiar with?\\
4. What is the person’s first language?\\
5. Does the person understand English?\\
\\
\textbf{Corresponding Answers}: "English"\\
\\
\textbf{Output}: \colorbox{red!10}{{[}\textit{LLM output}{]}} \\
\bottomrule
\end{tabular}
}
\caption{Prompt for question selection in automatic quality control.}
\label{tab:selectprompt}
\end{table*}

\paragraph{Question Selection.}
For question selection, we provide LLM with all the questions generated from previous step.
The quality control protocals are included in the instructions, as shown in Table~\ref{tab:selectprompt}.

\section{Experiment Implementation Details}

\subsection{In-Context Learning Prompt}
\label{app:iclprompt}

The ICL prompt consists of two parts.
First, we give the task description in the instruction.
Then, we provide $4$ demonstration examples.
The overall prompts are shown in Table~\ref{tab:baselineprompt}.

\begin{table*}[t]
\centering
\scalebox{0.85}{
\begin{tabular}{p{0.95\linewidth}ccccc}
\toprule
Prompt for in-context learning. \\
\midrule
\textbf{Instruction}: you are given one document and one anonymized real-world entity with one or more mentions in the passage. Then we will ask your a question about this anonymized entity. The questions cannot be answered solely within the document or the background knowledge. Your task is to leverage world knowledge you have like Wikipedia or wikidata as background knowledge combined with the given document to answer the question related to the anonymized entity. You must output all answers in the end. \\
 \\
\textbf{Document}:"[TV show A]" is the third episode of the first season of the American comedy television series The Office. Written by Paul Lieberstein, who also acts in the show as Toby Flenderson, and directed by Ken Whittingham, the episode first aired in the United States on April 5, 2005 on NBC. In this episode, Michael (Steve Carell) is tasked with choosing a new and inexpensive health care plan. He immediately hands it off to enthusiastic volunteer Dwight (Rainn Wilson). Dwight ruthlessly cuts nearly all benefits in the new plan, angering the rest of the office staff. Meanwhile, Pam (Jenna Fischer) and Jim (John Krasinski) make up fake diseases, much to Dwight's chagrin. In an attempt to appease them, Michael promises the entire office a surprise and then spends the rest of the day scrambling to come through with his promise. The employees wait for Michael's surprise, which he awkwardly never delivers. Jenna Fischer later called "[TV show A]" her favorite season one episode. During one particular scene, Rainn Wilson kept improvising new fake diseases. The laughter that resulted in his ad-libs was not scripted, as they were in fact the cast's genuine reaction to Wilson's fake diseases. The episode received a 2.9/7 in the Nielsen ratings among people aged 18–49 garnered 5.8 million viewers overall. In addition, the episode retained 100 \% of its lead - in 18–49 audience and ranked, along with the other first - season episodes of The Office, as NBC's highest - rated Tuesday night program since February 1, 2005. The episode received positive reviews. \\
\textbf{Question}: What is the series of TV show A?
\textbf{Answer}: "The Office" <stop>  \\
\\
\midrule
\textit{Here we omit other examples for better viewing.} \\
\midrule
\\
\textbf{Document}: "Insane" is the twelfth episode of the third season of the American animated sitcom [TV show A]. It originally aired on the Fox network in the United States on April 8, 2001. The episode was written by Bill Odenkirk and directed by Peter Avanzino. In the episode, Fry and Bender are admitted to an insane asylum for robots after being charged for their roles in holding up a bank. Fry's attempts to convince the asylum's staff that he is a human fail; he is eventually made to believe that he is a robot, and is deemed "cured" and released from the asylum. After being released, the Planet Express crew try to make him rediscover his humanity; these attempts fail, until Fry bleeds and realizes he is in fact, human. The episode introduces the recurring [TV show A] character Roberto. \\
\textbf{Question}: What is the publisher of TV show A? \\
\textbf{Answer}: \colorbox{red!10}{{[}\textit{LLM output}{]}} \\
\bottomrule
\end{tabular}
}
\caption{Prompt for question selection in automatic quality control.}
\label{tab:baselineprompt}
\end{table*}

\section{Supplementary Experiments}
\label{app:moreresult}

We evaluate our baseline models on \korct, \korch, and \korcl.
The results are shown in Table~\ref{tab:morebaseline}, as a supplementation to Table~\ref{tab:baseline}.

We observe that \korct, as a template-generated dataset, is the simplest among all the three versions.
Baselines generally achieve higher performance on \korct compared to \korch and even \korcl.
We also find that LLMs failed to successfully answer questions generated by themselves on \korcl.
This is because the questions are generated according to external knowledge guidance beyond LLM itself.

\begin{table}[!th]
\centering
\scalebox{0.84}{
\begin{threeparttable}
\setlength{\tabcolsep}{3.2pt}
\begin{tabular}{lccccccc}
\toprule
\multirow{2}{*}{\textbf{\korct}} & \multicolumn{3}{c}{P-ACC} & \multicolumn{3}{c}{P-F1} \\
\cmidrule(r){2-4}\cmidrule(lr){5-7}
& ID & OOD & Mean & ID & OOD & Mean \\
\midrule
BART-base  
& $55.8$ & $25.6$ & $45.2$
& $58.3$ & $30.9$ & $48.7$ \\
Flan-T5-base 
& $40.1$ & $25.8$ & $35.1$
& $42.4$ & $29.6$ & $37.9$ \\
\midrule
GPT-3 
& $17.3$ & $24.8$ & $19.9$
& $21.2$ & $30.6$ & $24.5$ \\
GLM-130B
& $9.0$ & $16.8$ & $11.7$
& $11.5$ & $20.5$ & $14.7$ \\
\midrule
RAG-seq 
& $60.6$ & $\mathbf{26.7}$ & $48.7$ 
& $62.1$ & $\mathbf{31.2}$ & $51.3$ \\
RAG-token 
& $64.0$ & $24.2$ & $50.0$
& $65.9$ & $28.4$ & $52.7$ \\
\midrule
EmbedKGQA 
& $\mathbf{66.7}$ & $22.9$ & $\mathbf{51.3}$
& $\mathbf{73.7}$ & $30.2$ & $\mathbf{58.5}$ \\
EmbedKGQA$^*$
& $39.9$ & $15.5$ & $31.3$
& $46.8$ & $23.4$ & $38.6$ \\
TransferNet 
& 35.8 & 14.9 & 28.5
& 40.7 & 19.2 & 33.2 \\
\bottomrule
\toprule
\multirow{2}{*}{\textbf{\korch}} & \multicolumn{3}{c}{P-ACC} & \multicolumn{3}{c}{P-F1} \\
\cmidrule(r){2-4}\cmidrule(lr){5-7}
& ID & OOD & Mean & ID & OOD & Mean \\
BART-base
& $50.3$ & $24.9$ & $41.4$
& $52.9$ & $30.2$ & $44.9$ \\
Flan-T5-base 
& $33.5$ & $24.0$ & $30.2$
& $35.8$ & $27.5$ & $32.9$ \\
\midrule
GPT-3 
& $18.2$ & $24.6$ & $20.5$
& $22.2$ & $30.2$ & $25.0$ \\
GLM-130B 
& $9.9$ & $14.9$ & $11.6$
& $12.7$ & $18.8$ & $14.8$ \\
\midrule
RAG-seq
& $\mathbf{61.7}$ & $\mathbf{25.9}$ & $\mathbf{49.2}$
& $63.7$ & $\mathbf{30.0}$ & $51.9$ \\
RAG-token 
& $57.4$ & $23.5$ & $45.5$
& $59.1$ & $27.2$ & $47.9$ \\
\midrule
EmbedKGQA 
& $61.2$ & $21.9$ & $47.4$ 
& $\mathbf{68.3}$ & $28.9$ & $\mathbf{54.5}$ \\
EmbedKGQA$^*$ 
& $34.0$ & $13.6$ & $26.9$
& $41.6$ & $21.8$ & $34.6$ \\
TransferNet 
& $32.7$ & $12.9$ & $25.8$
& $37.7$ & $16.6$ & $30.3$ \\
\bottomrule
\toprule
\multirow{2}{*}{\textbf{\korcl}} & \multicolumn{3}{c}{P-ACC} & \multicolumn{3}{c}{P-F1} \\
\cmidrule(r){2-4}\cmidrule(lr){5-7}
& ID & OOD & Mean & ID & OOD & Mean \\
\midrule
BART-base 
& $52.0$ & $27.9$ & $43.6$
& $54.7$ & $\mathbf{33.1}$ & $47.1$ \\
Flan-T5-base 
& $36.6$ & $26.6$ & $33.1$
& $38.9$ & $30.2$ & $35.8$ \\
\midrule
GPT-3
& $16.4$ & $24.1$ & $19.1$
& $20.5$ & $30.3$ & $23.9$ \\
GLM-130B 
& $9.2$ & $14.1$ & $10.9$
& $11.6$ & $17.9$ & $13.8$ \\
\midrule
RAG-seq 
& $\mathbf{64.8}$ & $\mathbf{28.7}$ & $\mathbf{52.2}$
& $66.7$ & $\mathbf{33.1}$ & $54.9$ \\
RAG-token 
& $56.8$ & $21.8$ & $44.5$
& $58.6$ & $25.7$ & $47.1$ \\
\midrule
EmbedKGQA 
& $62.7$ & $22.4$ & $48.6$
& $\mathbf{69.7}$ & $29.2$ & $\mathbf{55.5}$ \\
EmbedKGQA$^*$ 
& $42.8$ & $18.9$ & $34.4$
& $49.6$ & $26.0$ & $41.3$ \\
TransferNet 
& $31.8$ & $12.7$ & $25.1$
& $36.8$ & $16.2$ & $29.6$ \\
\bottomrule
\end{tabular}
\end{threeparttable}
}
\caption{
Baseline results on \korct, \korch, and \korcl.
EmbedKGQA$^*$ updates the knowledge representations during training, while EmbedKGQA uses freezed knowledge representations.
}
\label{tab:morebaseline}
\end{table}

\end{document}